\documentclass{article}

\usepackage[export]{adjustbox}
\usepackage{spconf_double,amsmath,graphicx}
\usepackage{multirow}
\usepackage{color}
\usepackage{rotating}
\usepackage{floatrow}
\usepackage{subcaption}
\twocolumn

\title{Personalizing Universal Recurrent Neural Network Language Model with User Characteristic Features by Social Network Crowdsourcing}
\name{Bo-Hsiang Tseng, Hung-yi Lee, and Lin-Shan Lee}
\address{National Taiwan Univeristy\\
		r02942037@ntu.edu.tw,   lslee@gate.sinica.edu.tw}
\begin{document}
%
\maketitle
\begin{abstract}
With the popularity of mobile devices, personalized speech recognizer becomes more realizable today and highly attractive. 
Each mobile device is primarily used by a single user, so it's possible to have a personalized recognizer well matching to the characteristics of individual user. 
Although acoustic model personalization has been investigated for decades, much less work have been reported on personalizing language model, probably because of the difficulties in collecting enough personalized corpora. 
Previous work used the corpora collected from social networks to solve the problem, but constructing a personalized model for each user is troublesome. 
In this paper, we propose a universal recurrent neural network language model with user characteristic features, so all users share the same model, except each with different user characteristic features. These user characteristic features can be obtained by crowdsouring over social networks, which include huge quantity of texts posted by users with known friend relationships, who may share some subject topics and wording patterns.
The preliminary experiments on Facebook corpus showed that this proposed approach not only drastically reduced the model perplexity, but offered very good improvement in recognition accuracy in n-best rescoring tests. 
This approach also mitigated the data sparseness problem for personalized language models.
\end{abstract}
\vspace{-0.5em}
\section{Introduction}
\label{sec:intro}
\vspace{-1em}

The personalization of various applications and services for each individual
user has been a major trend. Good examples include personalized web
search~\cite{c1,c2} and personalized recommendation systems~\cite{c3,c4,c5,c6}.
In the area of speech recognition, the popularity of mobile devices such as
smart phones and wearable clients makes personalized recognizers much more
realizable and highly attractive. Each mobile device is used primarily by a
single user, and can be connected to a personalized recognizer stored in the
cloud with much better performance, because this recognizer can be well-matched
to the linguistic characteristics of the individual user.

In acoustic model adaptation~\cite{c7,c8,c9},  personalization has been investigated for decades and has yielded very impressive
improvements with many approaches based on either HMM/GMM or
CD-DNN-HMM~\cite{CD-DNN-HMM}. However, there has been much less work reported on
language model (LM) personalization. LM adaptation has been studied for decades~\cite{old_adaptation_1,old_adaptation_2,old_adaptation_3}, but the previous works~\cite{LM_adaptation_1, LM_adaptation_2, LM_adaptation_3,LM_adaptation_4,LM_adaptation_5} primarily
focused on the problem of cross-domain or cross-genre linguistic mismatch, while
the cross-individual linguistic mismatch is often ignored. One good reason for
this is perhaps the difficulty in collecting personalized corpora for
personalized LMs.
However, this situation has changed in recent years. Nowadays, many
individuals post large quantities of texts over social networks, which yield
huge quantities of posted texts with known authors and given friend
relationships among the authors. It is therefore possible to
train personalized LMs because of the reasonable assumption that users with
close friend relationships may share common subject topics, wording habits,
and linguistic patterns.

Personalized LMs are useful in many aspects~\cite{personal_LM_1,personal_LM_2,personal_LM_3}.
In the area of speech recognition, personalization of LMs has been proposed and investigated for both N-gram-based
LMs~\cite{N-gram-based_personalization} and recurrent neural
networks (RNNLMs)~\cite{rnn-based_personalization} in the very limited previous
works. In these previous works, text posted by many individual users and
other information (such as friend relationships among users) were collected
from social networks. A background LM (either N-gram-based or RNN-based)
was then adapted toward an individual user's wording patterns by incorporating
social texts that the target user and other users had posted, considering
different aspects of their relationships and similarities between the users. In these
previous works, personalization was realized by training an LM for each
individual. There are inevitable shortcomings with this framework. First, even
with help of the social networks, it is not easy to obtain 
text corpora that are helpful for a particular user
for adapting a background LM towards a personalized LM. As a 
result, the personalized LM thus obtained easily overfits to the limited
data, and therefore yields relatively poor performance on the new data of the
target user. Second, to train and store a personalized LM for every user is in any case
time-consuming and memory-intensive, especially considering that the number of
users will only increase in the future.

Considering the above-mentioned defects in the previous framework, in this paper we
propose a new RNNLM-based paradigm for personalizing LMs. In
conventional RNNLM~\cite{conventional_RNNLM_1,conventional_RNNLM_2,conventional_RNNLM_3}, the 1-of-N encoding of each word is taken as the
input of the RNN, and then given the history word sequence, RNN outputs the
estimated probability distribution for the next word. In the new paradigm
proposed here, however, each user is represented by a feature vector encoding
some characteristics of the user, and this feature vector augments the 1-of-N
encoding feature of each word. A universal RNNLM is thus trained based on the data of these user
features,together with the texts over social networks by
a large number of users. The standard training method is used, except now the
same words produced by different users in the training set are augmented by
different user characteristic features. For each new user, his characteristic
feature is extracted to extend the 1-of-N word encoding, with which
the universal RNNLM can be used to recognize his speech. Because the same words
produced by different users are augmented with different features, given the
same history word sequence, the universal RNNLM can predict different
distributions of the next word for different users. In this way, the
personalization can be achieved even though all users share the same universal
RNNLM. This universal RNNLM trained from the social text produced by many users
is less liable to overfitting because a very large training set can be
obtained by aggregating the social texts of many users. Moreover, since the
recognizer for each user only requires the user's characteristic features rather
 than an entirely new model, the new paradigm saves time during training and 
memory in real-world implementations.
This concept of input features
for personalization is similar to the i-vectors used in deep neural network
(DNN) based acoustic models~\cite{dnn_1,dnn_2}, in which the i-vector of each
speaker is used to extend acoustic features such as MFCC. Preliminary
experiments show that the proposed method not only reduces model
perplexity but also reduces word error rates in n-best rescoring tests. In
addition, we find that this approach mitigates the overfitting problem for limited personalized data.
can be helpful in extracting the target user's characteristic features.

       \begin{figure}[!htb]
        \centering
        \includegraphics[width=\textwidth]{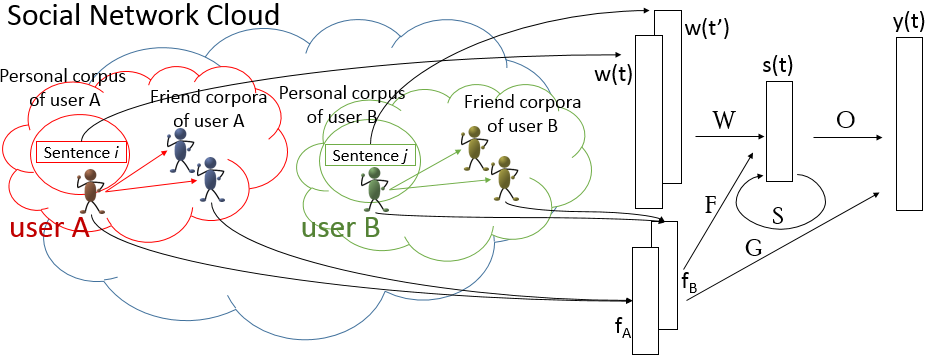}
		\vspace{-2em}
		  \caption{{\it The scenario for the proposed approach. When training
		  with sentence \textit{i} from user A, the user feature fed into the
		  RNNLM can be produced by either topic distribution of user A's personal corpus or searching over user A's personal / friends corpora
		  for sentences with topic
		  distributions closest to this sentence \textit{i}. }}
        \label{fig:crowd_rnn}
      \end{figure}

\vspace{-0.5em}
\section{LM Personalization Scenario}
\vspace{-0.5em}

Crowdsourcing~\cite{crowd_1,crowd_2} has varying definitions and has been applied
to a wide variety of tasks. For example, a crowdsourcing approach was proposed to
collect queries for information retrieval considering temporal
information~\cite{retrieval}. The MIT movie browser~\cite{browser_1,browser_2}
build a crowd-supervised spoken language
system. In this work, a cloud-based application was implemented offering users to 
access to their social network via voice, and was treated as a crowdsourcing platform
for collecting personal data. When the user logs into his Facebook account, he
may choose to grant this application the authority to collect his acoustic and
linguistic data for use in personalizing the voice access service. Users who do so
may enjoy the benefits of the superior recognition accuracy yielded by the
personalized recognizer via the crawled data.

Fig.~\ref{fig:crowd_rnn} depicts the scenario of the proposed approach. For
user A, the red figure in the left part of the figure,
the texts of his social network posts are
crawled to form the \textit{personal corpus} of the user (the red circle).
Besides, the posts of all user A's friends (blue figures) in social network are collected to form user A's \textit{friends corpora} 
(the red cloud surrounding the circle).
In previous work~\cite{rnn-based_personalization} the user's personal corpus and friends corpora was used for adapting a background LM trained with a large background corpus.
However, such LM adaptation
suffers from overfitting due to the limited adaptation data, and 
as mentioned above, also incurs heavy training/memory burdens.

In this paper, instead of building a personalized RNNLM for each user, a single
universal RNNLM is used by all users.
As shown in the right part of Fig.~\ref{fig:crowd_rnn}, a corpus of posts from a large group of users serves as the training data for the universal RNNLM.
This universal RNNLM comprises three layers: the input layer, the hidden layer, and the output layer,
very similar to those used previously\cite{conventional_RNNLM_1}, except the input layer 
is not only the word vector \textit{w(t)} 
representing the $t$-th word in a sentence using an 1-of-N encoding,
but concatenated with the additional user characteristic feature \textit{f}. 
This user characteristic feature is connected to both the hidden
layer \textit{s(t)} and output layer \textit{y(t)}~\footnote{This structure is
parallel to the context dependent RNNLM variant~\cite{conventional_RNNLM_3}
except that the context feature in the input layer is replaced by the user
characteristic feature \textit{f}.}.
This feature \textit{f} enables the model to take into account each specific user.
The network weights to be learned are the matrices \(W, F, S, G\) and \(O\) in
the right part of the figure.

\vspace{-0.5em}
\section{Extraction of user characteristic features}
\vspace{-0.5em}
We proposed two approaches to extract the user characteristic feature for each sentence, which are respectively described in Subsections~\ref{subsecsec:user} and~\ref{subsecsec:sentence}. 

\subsection{User-dependent Feature} \label{subsecsec:user}
\vspace{-0.5em}
In this approach, the personal corpus for each target user is viewed as a single document, and then a topic modeling approach is used to derive the topic distribution of that document. 
The topic distribution of the personal corpus thus represents the language characteristics of the user and is considered as the user characteristic feature {\it f} of the user. 
That is, during training the universal RNNLM, the 1-of-N encoding of the words in a personal corpus are all concatenated with the same topic distribution of that personal corpus.
The topic model used here is Latent Dirichlet Allocation (LDA)~\cite{LDA} model trained from a large corpus for many users. 

\vspace{-1em}
\subsection{Sentence-dependent Feature} \label{subsecsec:sentence}
\vspace{-0.5em}
Considering the fact that the personalized corpus of a user may cover many different topics, and the topic of the user may be switched dynamically and freely from one to another in the personal corpus, the topic distribution for the whole personal corpus may not very well represent each individual sentence within the personal corpus. 
On the other hand, even though the topic can be switched freely in the personal corpus of a user, we observe that it usually needs at least a few sentences to finish a specific topic. 
Therefore, to form a feature not only reflecting the characteristics of user but also a specific sentence, we can exploit a part  of the personal corpus whose topic distribution is close to the sentence. 
This may solve the problem of mismatch between the topic distribution of the whole personal corpus and each individual training sentence.

With the above consideration, in the second approach, every sentence in the personal corpus of a user has its unique feature {\it f} which is related to not only the user but the sentence itself.
In other words, the topic model is first used to infer the topic distribution of a sentence, we then use this topic distribution to search over the personal corpus of the user to find other N sentences whose topic distributions are most closest to one formed for the sentence being considered.
This search process is fast since it is limited to personal corpus of the considered user only.
During training the universal RNNLM, the average of the topic distributions of these N found sentences is taken as the user characteristic feature {\it f}, to be concatenated with the 1-of-N encoding features of the words in the sentence. 
Therefore, the same words in different sentences of a personal corpus may have different user characteristic features. 
We can also extend the search space to be over the friends corpora of the user as well.



The major difference of the two approaches in Subsections~\ref{subsecsec:user} and~\ref{subsecsec:sentence} lies in the concept of how a better language model can be obtained.
In the first approach, we assume the personal corpus of a user can reflect his language characteristics, so the data for inferring the topic distribution is the whole personal corpus. 
In the second approach, we assume a user actually switches his topic freely from sentence to sentence, so we try to find the similar sentences to construct the user character feature to reflect the language characteristics not only for the user but for the specific sentence itself.
So, the data to form the user characteristics is limited to the N sentences found in the search process. 
During testing, the user characteristic feature is obtained in exactly the same way, except the N-best list of an utterance was used with the LDA model to generate the topic distribution for an utterance.

\section{Effects of the user characteristic feature on RNNLM}
\vspace{-0.5em}

Here we use a real example from the Facebook data to show the effect of the user
characteristic features on RNNLM. 
User A left many posts about coffee in the
Facebook data, while user B never did so.
This yielded very different user characteristic features for the two users.
Here the user characteristic features mentioned in subsection~\ref{subsecsec:sentence} by searching for the N closest sentences are used.
Given the sentence \textit{``A bottle of milk can make 3 cups of latte''} which was
more likely to be produced by user A, we list in Table~\ref{tb:example} 
the perplexities evaluated by a
conventional RNNLM and the personalized RNNLM with different user
characteristic features.
The conventional RNNLM is in row (a). 
We see that the personalized RNNLM with the user characteristic feature
\textit{f$_A$} of user A produced a drastically decreased perplexity ( 152 vs 355, row (b) ) because of the well-matched characteristics,
while that with the user characteristic feature \textit{f$_B$} of user B
yielded a significantly increased perplexity ( 604 vs 355, row (c) ).

\begin{table}[h]
\centering
\begin{tabular}{|l|c|}
\hline
\multicolumn{1}{|c|}{Language Models} & Perplexity \\ \hline
(a) RNNLM (conventional)              & 355        \\ \hline
(b) RNNLM (with \textit{f$_A$})                   & 152        \\
(c) RNNLM (with \textit{f$_B$})                   & 604        \\ \hline
\end{tabular}
\vspace{-0.5em}
\caption{
The perplexity for sentence \textit{``A bottle of milk can make 3 cups of
latte''} using different models, where user A's personal corpus included 
many posts about coffee but user B's personal corpus contained no such posts.
}
\label{tb:example}
\end{table}

   \vspace{-1.5em}
   \section{Experimental Setup}
   \vspace{-0.5em}
    \subsection{Corpus \& LMs}
    \vspace{-0.5em}
Our experiments were conducted on a crawled Facebook corpus.
A total of 42 users logged in and authorized this project to collect for research their posts and basic information.
These 42 users were our target users, and were divided into 3 groups for cross validation, i.e., to train the universal LM using the data of two groups and test those for the rest. 
Furthermore, with their consent, the observable public data (the personal and friends corpora) of these 42 target users were also available for the experiments.
This resulted in the personal data of 93,000 anonymous people and a total of 2.4 million sentences.
The number of sentences for each user among the 93,000 ranged from 1 to 8,566 with a mean of 25.7, comprising 10.6 words (Chinese, English, or mixed) per sentence on average.
A total of 12,000 sentences for the 42 target users was taken as the testing set, and among them 948 produced by the respective target users were taken as testing utterances for ASR experiments.

For the background corpus, 500k sentences were collected from the popular
social networking site Plurk to train the topic model.
Using the Mallet toolkit~\cite{Mallet}, 
we trained a latent Dirichlet allocation-based (LDA) topic model,
taking each sentence as a document.
The modified Kneser-Ney algorithm~\cite{m_kn} was used for the N-gram LM
smoothing.
From the corpus the most frequent 18,000 English words and 46,000 Chinese words were selected to form the lexicon.
The SRILM~\cite{srilm} toolkit was used for the N-gram LM training and adaptation,
while RNNLM toolkit~\cite{rnn_toolkit} was used for RNNLM here.

\vspace{-0.5em}
\subsection{N-best rescoring}
\vspace{-0.5em}
To generate the 1,000-best lists for rescoring, 
we used lattices produced using the HTK toolkit~\cite{HTK}.
To generate the lattices we used a trigram LM
adapted to the personal and friend corpora using Kneser-Ney smoothing (KN3).
For first-pass decoding we used Mandarin triphone models
trained on the ASTMIC corpus and the English triphone
models trained on the Sinica Taiwan English corpus~\cite{ASTMIC}; both corpora include
hundreds of speakers. Both sets of models were adapted using unsupervised MLLR.

    \section{Experimental Results}
    \vspace{-0.5em}
    
    	\subsection{Extraction of user characteristic features}
\begin{figure}[htb]
    \centering
		 \includegraphics[scale=0.575]{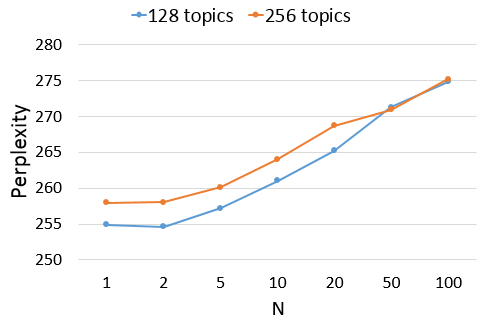}
		  \caption{{\it Perplexities for different number of LDA topics and different number of similar sentences ($N$) selected to build the user characteristic feature in~\ref{subsecsec:sentence}.}}
        \label{fig:fs}
    \end{figure}

        \vspace{-0.5em}
As mentioned in Section~\ref{subsecsec:sentence}, only those $N$ sentences most close to the
sentence under consideration were used to build the user characteristic feature.
Fig.~\ref{fig:fs} are the perplexities for different $N$ 
(out of the user plus friend corpora) and different number of topics for 
LDA~\footnote{Only one-tenth of the personal and friend corpus was used in these
preliminary experiments.}.
The figure shows that there was almost no difference between $N=1$ and $N=2$, 
but as $N$ increased beyond 2 the perplexity also increased, suggesting a wide 
variety of topics even for the same user and his friends. 
We thus chose $N=1$ for the following experiments.
        
\vspace{-0.5em}
\subsection{Perplexity}
\vspace{-0.5em}
        
\begin{table}
\centering
\vspace{-0.5em}
\caption{
Perplexity (PPL) Results.
KN3 represents Kneser-Ney tri-gram, while \lq RNN/model\rq and \lq RNN/UCF\rq are for Personalizing RNNLM based on model adaption (model) and the user characteristic feature (UCF) approach proposed here respectively.
Notation \lq B\rq, \lq B+P\rq and \lq B+P+F\rq  respectively indicate using only background corpus (B), plus personal corpus (B+P), and plus friends corpora in addition (B+P+F).
Notation \lq UD\rq and \lq SD\rq respectively indicate extracting user-dependent and sentence-dependent features in the proposed approach.
The results for RNNLM with hidden layer size of 50 and 200 are listed.}
\label{tb:PPL}
\begin{tabular}{|l|l|c|c|}
\hline
\multicolumn{2}{|c|}{Perplexity}               & h50                      & h200      \\ \hline
\multirow{3}{*}{(a)} &(a-1) KN3, B                  & \multicolumn{2}{c|}{343}             \\
                     & (a-2) KN3, B+P                & \multicolumn{2}{c|}{299}             \\
                     & (a-3) KN3, B+P+F              & \multicolumn{2}{c|}{233}             \\ \hline
(b)                  & RNN, B                  & 441                      & 398       \\ \hline
\multirow{2}{*}{(c)} & (c-1) RNN/model, B+P          & \multicolumn{1}{l|}{350} & 320       \\
& (c-2) RNN/model, B+P+F        & 296                      & 265       \\
 \hline
\multirow{4}{*}{(d)} 
                     & (d-1) RNN/UCF, UD, B+P             & \multicolumn{1}{l|}{313} & 270       \\
                     & (d-2) RNN/UCF, SD, B+P        & \multicolumn{1}{l|}{269} & 218 \\ 
                     & (d-3) RNN/UCF, SD, B+P+F      & 229                      & 211       \\
                     & (d-4) RNN/UCF, SD, B+P+F, avg & 192                      & 165 \\ \hline
\end{tabular}
\end{table}

Table~\ref{tb:PPL} shows the results of perplexity (PPL). 
Personalized Kneser-Ney tri-gram is reported in section (a)~\cite{rnn-based_personalization}, where \lq B\rq, \lq B+P\rq, \lq B+P+F\rq  indicate respectively background (B), background plus personal corpus (B+P), and plus friends corpora in addition (B+P+F).
Row (b) is RNNLM using only background corpus (\lq B\rq) without any personalization with hidden layer size of 50 and 200.
Personalizing RNNLM based on model adaption (model)~\cite{rnn-based_personalization} and the user characteristic feature (UCF) approach proposed here are respectively labeled with \lq RNN/model\rq in section (c) and 'RNN/UCF' in section (d).
In section (d), notations \lq UD\rq and \lq SD\rq respectively indicate extracting user-dependent (subsection~\ref{subsecsec:user}) and sentence-dependent (subsection~\ref{subsecsec:sentence}) features in the proposed approach.

We found that sentence-dependent feature outperformed the user-dependent feature ( (d-2) v.s. (d-1) ), which implies that the considerations of topic switching in subsection~\ref{subsecsec:sentence} is reasonable.
Under the condition involving personal corpora (B+P), no matter using user-dependent  or sentence-dependent  features, the approach proposed here is always better than the model adaption approach ( (d-1) (d-2) v.s. (c-1) ).
With the sentence-dependent features, PPL improvement is up to 102 ( 218 in (d-2) v.s. 320 in (c-1) in h200 ). 
This means extracting a good feature to characterize the user is more efficient than using personal data to learn a personalized RNNLM.
With the friends corpora involved (B+P+F)\footnote{When extracting the sentence-dependent feature, search space is over both personal and friends corpora.}, the proposed approach is still better than the model adaption approach ( 211 in (d-3) v.s. 265 in (c-2) ).
When using sentence-dependent feature we may further average the found user characteristic feature with the topic distribution of the sentence being considered ( RNN/UCF, SD, B+P+F, avg in (d-4) ).
PPL in this case can be further improved (165 in ( d-4) v.s. 211 in (d-3) ).
With this best model obtained here ( RNN/UCF, SD, B+P+F, avg ), the perplexity is reduced by 58.5\% compared to RNNLM without personalization ( RNN, B in (b) ), 37.7\% compared to the model adaption approach with friends corpora used ( RNN/model, B+P+F in (c-2) ). 

\begin{table}[ht]
\centering
\vspace{-0.5em}
\caption{
Word error rate (WER) results with same notations as in Table~\ref{tb:PPL}.
For sentence-dependent (SD) features, the topic distributions are estimated from N-best lists in section (d), while from reference transcriptions in section (e) (oracle).}
\label{tb:WER}
\begin{tabular}{|l|l|c|c|}
\hline
\multicolumn{2}{|c|}{WER (\%)}                                                                  & h50                        & h200                             \\ \hline
\multirow{3}{*}{(a)}                                                  & KN3, B                  & \multicolumn{2}{c|}{43.80}                                    \\
                                                                      & (a-1) KN3, B+P                & \multicolumn{2}{c|}{43.39}                                    \\
                                                                      & (a-2) KN3, B+P+F              & \multicolumn{2}{c|}{41.95}                                    \\ \hline
(b)                                                                   & (a-3) RNN, B                  & 41.12                      & 41.14                            \\ \hline
\multirow{2}{*}{(c)}                                                  & (c-1) RNN/model, B+P          & 40.84                      & 40.87                            \\
                                                                      & (c-2) RNN/model, B+P+F        & 40.71                      & 40.68                            \\ \hline
\multirow{4}{*}{(d)}                                                  & (d-1) RNN/UCF, UD, B+P             & 40.48                      & 40.16                      \\
                                                                      & (d-2) RNN/UCF, SD, B+P        & 40.47                      & 40.36                            \\
                                                                      & (d-3) RNN/UCF, SD, B+P+F      & 40.43                      & 40.40                            \\
                                                                      & (d-4) RNN/UCF, SD, B+P+F, avg & 40.23                      &  40.26                    \\ \hline
\multirow{4}{*}{\begin{tabular}[c]{@{}l@{}}(e)\\ {\rotatebox{90}{\kern-0.5em oracle}}\end{tabular}} & (e-1) RNN/UCF, UD, B+P             & \multicolumn{1}{l|}{40.48} & \multicolumn{1}{l|}{40.16} \\
                                                                      & (e-2) RNN/UCF, SD, B+P        & 40.15                      & 40.09                            \\
                                                                      & (e-3) RNN/UCF, SD, B+P+F      & 40.03                      & 39.95                            \\
                                                                      & (e-4) RNN/UCF, SD, B+P+F, avg & 39.40                      &  39.45                      \\ \hline
\end{tabular}
\end{table}

\vspace{-0.5em}
\subsection{Word error rate (WER)}
\vspace{-0.5em}

Table~\ref{tb:WER} reports the word error rates (WER) with the same notation as in Table~\ref{tb:PPL}.
Section (a) is for the three different tri-gram LMs without and with personalization. 
As expected, with more adaptation data, the tri-gram LMs performed better ( (a-3)\textless(a-2)\textless(a-1) )~\cite{rnn-based_personalization}.
We used the best adapted tri-gram LM ( KN3,B+S+F in (a-3) ) to generate 1000-best lists for RNNLM rescoring.
Section (b) is for rescoring results using RNNLM without personalization, while sections (c) and (d) are for model adaption (model) approach and the proposed approach (UCF) respectively.
For sentence-dependent (SD) features, we viewed the 1000-best list as a single document and used the LDA topic model to infer the topic distribution, and then search for the closest sentences as mentioned in subsection~\ref{subsecsec:sentence} to construct the sentence-dependent feature of each utterance for rescoring.
For user-dependent (UD) features in the proposed approach, because the feature is extracted from the personal corpus of the user and independent of the input utterance, so the feature extraction process does not depend on ASR.
Regardless of the features used and the data involved, the proposed approach was always better than the model adaption approach ( (d-1, 2) v.s. (c-1) and (d-3, 4) v.s. (c-2) ).

To our surprise, for 200 hidden layer units the proposed approach with user-dependent feature is better than sentence-dependent feature in terms of WER ( (d-1) v.s. (d-2, 3, 4) for h200 ). 
This may be because the user-dependent feature is estimated from the training corpus of target user thus not influenced by ASR errors at all; while for sentence-dependent feature, the topic distribution from N-best list was inaccurate due to ASR errors.
This was verified  in the oracle experiments in section (e), in which we used the topic distribution of the reference transcription of the utterance to replace the topic distribution of N-best list and do the rescoring\footnote{In the oracle experiments in section (e), results of user-dependent feature (e-1) were the same as those in (d-1) because ASR was not involved in extracting the feature.}.
Here we see the sentence-dependent feature (SD) is better than user-dependent feature (UD) ( (e-2) v.s. (e-1) while (e-3) (e-4) are even better ).
Also, with topic distributions from the reference transcriptions, the results of sentence-dependent feature can be improved by absolute 0.81\% ( from 40.26\% in (d-4) to 39.45\% in (e-4) for h200 ).

So for the real best result here ( RNN/UCF, UD, B+P for h200 in (d-1) ), we reduced WER by 1.79\% compared to the best of KN3 including friends corpora ( 41.95\% in (a-3) ), from which the 1000-best lists were obtained,  0.98\% compared to RNNLM without personalization ( 41.14\% in (b) ), 0.52\% compared to the best of model adaption approach ( 40.68\% in c-2) ). 
In the oracle case the best result can be even much better ( 39.45\% RNN/UCF, SD, B+P+F, avg in (e-4) ), which indicates the space for further improvement.
    
      \begin{figure*}[htb]
        \centering
        \includegraphics[scale=0.65]{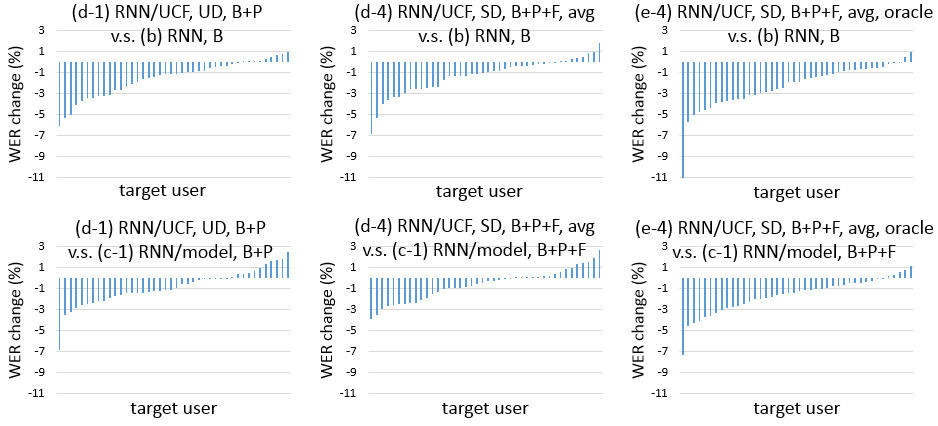}
        
		  \caption{WER changes across all 42 target users. The three figures in the upper row compare respectively the proposed approach with user-dependent feature ( RNN/UCF, UD, B+P, in (d-1) of Table~\ref{tb:WER} ), sentence-dependent feature ( RNN/UCF, SD, B+P+F, avg, in (d-4 ) of Table~\ref{tb:WER}) and sentence-dependent feature in oracle experiments ( RNN/UCF, SD, B+P+F, avg, in (e-4) ) with the baseline of RNNLM without personalization ( row (b) in Table~\ref{tb:WER} ), and the three figures in lower row compare the same three approaches with the model adaption approach ( row (c-1) in Table~\ref{tb:WER} ).}
        \label{fig:WER_Difference}
      \end{figure*}

\subsection{Analysis}
\subsubsection{WER over all target users}
\vspace{-0.5em}
Because the average didn't tell whether the proposed approach is actually helpful for most users or for just very few users, we plot in addition the WER change obtained across the all 42 target users in Fig.~\ref{fig:WER_Difference}.
The three figures in the upper row compare respectively the proposed approach with user-dependent feature ( RNN/UCF, UD, B+P, in (d-1) of Table~\ref{tb:WER} ), sentence-dependent feature ( RNN/UCF, SD, B+P+F, avg, in (d-4) of Table~\ref{tb:WER} ) and sentence-dependent feature in oracle experiments ( RNN/UCF, SD, B+P+F, avg, in (e-4) ) with the baseline of RNNLM without personalization ( row (b) in Table~\ref{tb:WER} ), and the three figures in lower row compare the same three approaches with the model adaption approach ( row (c-1) in Table~\ref{tb:WER} ).
Each figure has 42 bars for the 42 target users, sorted based on the WER change.
Here a negative value means that the proposed approach here offered WER reduction to the user.
From Fig.~\ref{fig:WER_Difference}, we see the proposed approach offered better performance to most target users.
For example, in the first figure on the upper row ( (d-1) RNN/UCF, UD, B+P v.s. (b) RNN,B ), 9 users had worse WER with our approach, all by less than 1\%, but all other users had WER reduction, 24 of them by more than 1\%.
Similar for the rest cases.
The results show the proposed approach offered improvements to most target users.


\vspace{-0.5em}
\subsubsection{Size of personal corpus}
\begin{figure}[htb]
        \centering
        \includegraphics[scale=0.455]{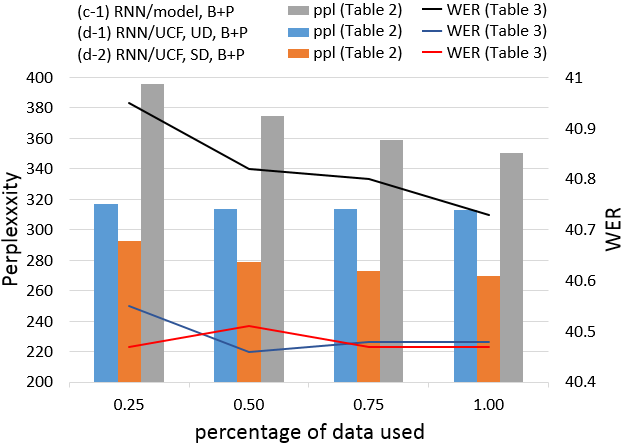}
        \caption{{\it 
Perplexity (PPL) and word error rate (WER) for different sizes of personal corpus for the model adaptation approach ( (c-1) RNN/model, B+P) and the proposed approaches ( (d-1) RNN/UCF, UD, B+P and (d-2) RNN/UCF, SD, B+P ).
The horizontal axis is the percentage of the original personal corpora used
( 1.00 on the right is for the data in Tables~\ref{tb:PPL} and~\ref{tb:WER} ).
        }}
        \label{fig:Usage of data}
    \end{figure}

\vspace{-0.5em}
As mentioned above, the model adaptation approach results in overfitting to the limited personal data
and may yield poor performance on a particular user's new data.
This is illustrated in Fig.~\ref{fig:Usage of data}.
The horizontal axis of the figure is the percentage of the
original personal corpus used, where 1.00 means using the entire original personal corpus, that is, those cases (c-1) RNN/model, (d-1) RNN/UCF, UD, B+P and (d-2) RNN/UCF, SD, B+P in Tables~\ref{tb:PPL} and~\ref{tb:WER} for h50.
We see that as less data were available, the proposed approach (d-1) and (d-2) demonstrated much smaller increases in perplexity and  much more stable WER, whereas for the model adaptation approach (c-1), the perplexity and WER increased significantly at a greater rate.
The result of different size of friends corpora has the same trend.

\section{Conclusions}
\vspace{-0.5em}
In this paper, we proposed a new framework for personalizing a universal RNNLM
using data crawled over social networks. 
The proposed approach is based on a user characteristic feature extracted
from the user corpus and friends corpora, which is not only user-dependent but sentence-dependent feature.
This universal RNNLM can predict different word distributions for different
users given the same context.
Experiments demonstrated really good improvements in both perplexity
and WER, and the proposed approach is much more robust to data
sparseness than the previous work.

\newpage
\bibliographystyle{IEEEbib}
\bibliography{strings,refs}

\begin{thebibliography}{10}

\bibitem{c1}
Geoffrey Zweig and Shuangyu Chang,
\newblock ``Personalizing model m for voice-search.,''
\newblock in {\em INTERSPEECH}, 2011, pp. 609--612.

\bibitem{c2}
Micro Speretta and Susan Gauch,
\newblock ``Personalized search based on user search histories,''
\newblock in {\em Web Intelligence, 2005. Proceedings. The 2005 IEEE/WIC/ACM
  International Conference on}. IEEE, 2005, pp. 622--628.

\bibitem{c3}
Yoon~Ho Cho, Jae~Kyeong Kim, and Soung~Hie Kim,
\newblock ``A personalized recommender system based on web usage mining and
  decision tree induction,''
\newblock {\em Expert systems with Applications}, vol. 23, no. 3, pp. 329--342,
  2002.

\bibitem{c4}
Yehuda Koren, Robert Bell, and Chris Volinsky,
\newblock ``Matrix factorization techniques for recommender systems,''
\newblock {\em Computer}, , no. 8, pp. 30--37, 2009.

\bibitem{c5}
Frank~Edward Walter, Stefano Battiston, and Frank Schweitzer,
\newblock ``A model of a trust-based recommendation system on a social
  network,''
\newblock {\em Autonomous Agents and Multi-Agent Systems}, vol. 16, no. 1, pp.
  57--74, 2008.

\bibitem{c6}
Moon-Hee Park, Jin-Hyuk Hong, and Sung-Bae Cho,
\newblock ``Location-based recommendation system using bayesian user’s
  preference model in mobile devices,''
\newblock in {\em Ubiquitous Intelligence and Computing}, pp. 1130--1139.
  Springer, 2007.

\bibitem{c7}
Christopher~J Leggetter and Philip~C Woodland,
\newblock ``Maximum likelihood linear regression for speaker adaptation of
  continuous density hidden markov models,''
\newblock {\em Computer Speech \& Language}, vol. 9, no. 2, pp. 171--185, 1995.

\bibitem{c8}
Phil~C Woodland,
\newblock ``Speaker adaptation for continuous density hmms: A review,''
\newblock in {\em ISCA Tutorial and Research Workshop (ITRW) on Adaptation
  Methods for Speech Recognition}, 2001.

\bibitem{c9}
Jean-Luc Gauvain and Chin-Hui Lee,
\newblock ``Maximum a posteriori estimation for multivariate gaussian mixture
  observations of markov chains,''
\newblock {\em Speech and audio processing, ieee transactions on}, vol. 2, no.
  2, pp. 291--298, 1994.

\bibitem{CD-DNN-HMM}
Geoffrey Hinton, Li~Deng, Dong Yu, George~E Dahl, Abdel-rahman Mohamed, Navdeep
  Jaitly, Andrew Senior, Vincent Vanhoucke, Patrick Nguyen, Tara~N Sainath,
  et~al.,
\newblock ``Deep neural networks for acoustic modeling in speech recognition:
  The shared views of four research groups,''
\newblock {\em Signal Processing Magazine, IEEE}, vol. 29, no. 6, pp. 82--97,
  2012.

\bibitem{old_adaptation_1}
Roland Kuhn and Renato De~Mori,
\newblock ``A cache-based natural language model for speech recognition,''
\newblock {\em Pattern Analysis and Machine Intelligence, IEEE Transactions
  on}, vol. 12, no. 6, pp. 570--583, 1990.

\bibitem{old_adaptation_2}
Rukmini~M Iyer and Mari Ostendorf,
\newblock ``Modeling long distance dependence in language: Topic mixtures
  versus dynamic cache models,''
\newblock {\em Speech and Audio Processing, IEEE Transactions on}, vol. 7, no.
  1, pp. 30--39, 1999.

\bibitem{old_adaptation_3}
Ronald Rosenfeld,
\newblock ``A maximum entropy approach to adaptive statistical language
  modelling,''
\newblock {\em Computer Speech \& Language}, vol. 10, no. 3, pp. 187--228,
  1996.

\bibitem{LM_adaptation_1}
Jerome~R Bellegarda,
\newblock ``Statistical language model adaptation: review and perspectives,''
\newblock {\em Speech communication}, vol. 42, no. 1, pp. 93--108, 2004.

\bibitem{LM_adaptation_2}
Aaron Heidel and Lin-shan Lee,
\newblock ``Robust topic inference for latent semantic language model
  adaptation,''
\newblock in {\em Automatic Speech Recognition \& Understanding, 2007. ASRU.
  IEEE Workshop on}. IEEE, 2007, pp. 177--182.

\bibitem{LM_adaptation_3}
Bo-June~Paul Hsu and James Glass,
\newblock ``Style \& topic language model adaptation using hmm-lda,''
\newblock in {\em Proceedings of the 2006 Conference on Empirical Methods in
  Natural Language Processing}. Association for Computational Linguistics,
  2006, pp. 373--381.

\bibitem{LM_adaptation_4}
David Mrva and Philip~C Woodland,
\newblock ``Unsupervised language model adaptation for mandarin broadcast
  conversation transcription.,''
\newblock in {\em INTERSPEECH}. Citeseer, 2006.

\bibitem{LM_adaptation_5}
Yik-Cheung Tam and Tanja Schultz,
\newblock ``Correlated latent semantic model for unsupervised lm adaptation,''
\newblock in {\em Acoustics, Speech and Signal Processing, 2007. ICASSP 2007.
  IEEE International Conference on}. IEEE, 2007, vol.~4, pp. IV--41.

\bibitem{personal_LM_1}
Yu-Yang Huang, Rui Yan, Tsung-Ting Kuo, and Shou-De Lin,
\newblock ``Enriching cold start personalized language model using social
  network information,''
\newblock {\em ACL’14}, pp. 611--617, 2014.

\bibitem{personal_LM_2}
Gui-Rong Xue, Jie Han, Yong Yu, and Qiang Yang,
\newblock ``User language model for collaborative personalized search,''
\newblock {\em ACM Transactions on Information Systems (TOIS)}, vol. 27, no. 2,
  pp. 11, 2009.

\bibitem{personal_LM_3}
Arjumand Younus, Colm O’Riordan, and Gabriella Pasi,
\newblock ``A language modeling approach to personalized search based on
  users’ microblog behavior,''
\newblock in {\em Advances in Information Retrieval}, pp. 727--732. Springer,
  2014.

\bibitem{N-gram-based_personalization}
Tsung-Hsien Wen, Hung-Yi Lee, Tai-Yuan Chen, and Lin-Shan Lee,
\newblock ``Personalized language modeling by crowd sourcing with social
  network data for voice access of cloud applications,''
\newblock in {\em Spoken Language Technology Workshop (SLT), 2012 IEEE}. IEEE,
  2012, pp. 188--193.

\bibitem{rnn-based_personalization}
Tsung-Hsien Wen, Aaron Heidel, Hung-yi Lee, Yu~Tsao, and Lin-Shan Lee,
\newblock ``Recurrent neural network based language model personalization by
  social network crowdsourcing.,''
\newblock in {\em INTERSPEECH}, 2013, pp. 2703--2707.

\bibitem{conventional_RNNLM_1}
Tomas Mikolov, Martin Karafi{\'a}t, Lukas Burget, Jan Cernock{\`y}, and Sanjeev
  Khudanpur,
\newblock ``Recurrent neural network based language model.,''
\newblock in {\em INTERSPEECH 2010, 11th Annual Conference of the International
  Speech Communication Association, Makuhari, Chiba, Japan, September 26-30,
  2010}, 2010, pp. 1045--1048.

\bibitem{conventional_RNNLM_2}
Tom{\'a}{\v{s}} Mikolov, Stefan Kombrink, Luk{\'a}{\v{s}} Burget, Jan~Honza
  {\v{C}}ernock{\`y}, and Sanjeev Khudanpur,
\newblock ``Extensions of recurrent neural network language model,''
\newblock in {\em Acoustics, Speech and Signal Processing (ICASSP), 2011 IEEE
  International Conference on}. IEEE, 2011, pp. 5528--5531.

\bibitem{conventional_RNNLM_3}
Tomas Mikolov and Geoffrey Zweig,
\newblock ``Context dependent recurrent neural network language model.,''
\newblock in {\em SLT}, 2012, pp. 234--239.

\bibitem{dnn_1}
George Saon, Hagen Soltau, David Nahamoo, and Michael Picheny,
\newblock ``Speaker adaptation of neural network acoustic models using
  i-vectors,''
\newblock in {\em Automatic Speech Recognition and Understanding (ASRU), 2013
  IEEE Workshop on}. IEEE, 2013, pp. 55--59.

\bibitem{dnn_2}
Vishwa Gupta, Patrick Kenny, Pierre Ouellet, and Themos Stafylakis,
\newblock ``I-vector-based speaker adaptation of deep neural networks for
  french broadcast audio transcription,''
\newblock in {\em Acoustics, Speech and Signal Processing (ICASSP), 2014 IEEE
  International Conference on}. IEEE, 2014, pp. 6334--6338.

\bibitem{crowd_1}
Anhai Doan, Raghu Ramakrishnan, and Alon~Y Halevy,
\newblock ``Crowdsourcing systems on the world-wide web,''
\newblock {\em Communications of the ACM}, vol. 54, no. 4, pp. 86--96, 2011.

\bibitem{crowd_2}
Robert Munro, Steven Bethard, Victor Kuperman, Vicky~Tzuyin Lai, Robin Melnick,
  Christopher Potts, Tyler Schnoebelen, and Harry Tily,
\newblock ``Crowdsourcing and language studies: the new generation of
  linguistic data,''
\newblock in {\em Proceedings of the NAACL HLT 2010 Workshop on Creating Speech
  and Language Data with Amazon's Mechanical Turk}. Association for
  Computational Linguistics, 2010, pp. 122--130.

\bibitem{retrieval}
Klaus Berberich, Srikanta Bedathur, Omar Alonso, and Gerhard Weikum,
\newblock {\em A language modeling approach for temporal information needs},
\newblock Springer, 2010.

\bibitem{browser_1}
Jingjing Liu, Scott Cyphers, Panupong Pasupat, Ian McGraw, and Jim Glass,
\newblock ``A conversational movie search system based on conditional random
  fields.,''
\newblock in {\em INTERSPEECH}, 2012.

\bibitem{browser_2}
Ian McGraw, Scott Cyphers, Panupong Pasupat, Jingjing Liu, and Jim Glass,
\newblock ``Automating crowd-supervised learning for spoken language
  systems.,''
\newblock in {\em INTERSPEECH}, 2012.

\bibitem{LDA}
David~M Blei, Andrew~Y Ng, and Michael~I Jordan,
\newblock ``Latent dirichlet allocation,''
\newblock {\em the Journal of machine Learning research}, vol. 3, pp.
  993--1022, 2003.

\bibitem{Mallet}
Andrew~Kachites McCallum,
\newblock ``Mallet: A machine learning for language toolkit,''
\newblock http://mallet.cs.umass.edu, 2002.

\bibitem{m_kn}
Frankie James,
\newblock ``Modified kneser-ney smoothing of n-gram models,''
\newblock {\em Research Institute for Advanced Computer Science, Tech. Rep.
  00.07}, 2000.

\bibitem{srilm}
Andreas Stolcke et~al.,
\newblock ``Srilm-an extensible language modeling toolkit.,''
\newblock in {\em INTERSPEECH}, 2002.

\bibitem{rnn_toolkit}
Tomas Mikolov, Stefan Kombrink, Anoop Deoras, Lukar Burget, and Jan Cernocky,
\newblock ``Rnnlm-recurrent neural network language modeling toolkit,''
\newblock in {\em Proc. of the 2011 ASRU Workshop}, 2011, pp. 196--201.

\bibitem{HTK}
Steve Young, Gunnar Evermann, Mark Gales, Thomas Hain, Dan Kershaw, Xunying
  Liu, Gareth Moore, Julian Odell, Dave Ollason, Dan Povey, et~al.,
\newblock {\em The HTK book}, vol.~2,
\newblock Entropic Cambridge Research Laboratory Cambridge, 1997.

\bibitem{ASTMIC}
Ching-Feng Yeh, Aaron Heidel, Hong-Yi Lee, and Lin-Shan Lee,
\newblock ``Recognition of highly imbalanced code-mixed bilingual speech with
  frame-level language detection based on blurred posteriorgram,''
\newblock in {\em Acoustics, Speech and Signal Processing (ICASSP), 2012 IEEE
  International Conference on}. IEEE, 2012, pp. 4873--4876.

\end{thebibliography}

\end{document}